\documentclass[runningheads]{llncs}
\usepackage{graphicx}
\usepackage{amsmath,amssymb}
\usepackage[footnotesize]{subfig}
\usepackage{epsfig}
\usepackage{graphicx}
\usepackage{booktabs}
\usepackage{tabulary}
\usepackage{multirow}
\usepackage{url}
\usepackage{tablefootnote}
\usepackage[10pt]{moresize}
\usepackage[hang]{footmisc}
\usepackage{pifont}
\usepackage[colorlinks]{hyperref}
\usepackage{color}
\usepackage[letterspace=-93]{microtype}

\begin{document}
\pagestyle{headings}
\mainmatter
\def\ECCV18SubNumber{597}  

\title{Macro-Micro Adversarial Network \\ for Human Parsing} 

\titlerunning{Macro-Micro Adversarial Network for Human Parsing}

\author{Yawei Luo\inst{1,2} \and
Zhedong Zheng\inst{2} \and
Liang Zheng\inst{2,3} \and
Tao Guan\inst{1} \and
Junqing Yu\inst{1} \and
Yi Yang\inst{2}}

\authorrunning{Y. Luo et al.}
\institute{School of Computer Science and Technology, \\Huazhong University of Science and Technology \email{ \{royalvane,qd\_gt,yjqing\}@hust.edu.cn} \and
CAI, University of Technology Sydney\\ \and Singapore University of Technology and Design \email{ \{zdzheng12,liangzheng06,yee.i.yang\}@gmail.com}
 }

\maketitle
\begin{abstract}
In human parsing, the pixel-wise classification loss has drawbacks in its low-level local inconsistency and high-level semantic inconsistency. The introduction of the adversarial network tackles the two problems using a single discriminator. However, the two types of parsing inconsistency are generated by distinct mechanisms, so it is difficult for a single discriminator to solve them both. To address the two kinds of inconsistencies, this paper proposes the Macro-Micro Adversarial Net (MMAN). It has two discriminators. One discriminator, Macro $D$, acts on the low-resolution label map and penalizes semantic inconsistency, \emph{e.g.,} misplaced body parts. The other discriminator, Micro $D$, focuses on multiple patches of the high-resolution label map to address the local inconsistency, \emph{e.g.,} blur and hole. Compared with traditional adversarial networks, MMAN not only enforces local and semantic consistency explicitly, but also avoids the poor convergence problem of adversarial networks when handling high resolution images.
In our experiment, we validate that the two discriminators are complementary to each other in improving the human parsing accuracy. The proposed framework is capable of producing competitive parsing performance compared with the state-of-the-art methods, \emph{i.e.,} mIoU=46.81\% and 59.91\% on LIP and PASCAL-Person-Part, respectively. On a relatively small dataset PPSS, our pre-trained model demonstrates impressive generalization ability. The code is publicly available at \url{https://github.com/RoyalVane/MMAN}.

\end{abstract}

\section{Introduction}
Human parsing aims to segment a human image into multiple semantic parts. It is a pixel-level prediction task which requires to understand human images in both the global level and  the local level. Human parsing can be widely applied to human behavior analysis \cite{gan2016concepts}, pose estimation \cite{xia2016pose} and fashion synthesis \cite{iccv2017fashiongan}.
Recent advances in human parsing and semantic segmentation \cite{liang2015deep,xia2016pose,gong2017look,long2015fully,zhang2018adversarial,zhang2018self} mostly explore the potential of the convolutional neural network (CNN).

\begin{figure}[t]
\centering
\includegraphics[width=1\linewidth]{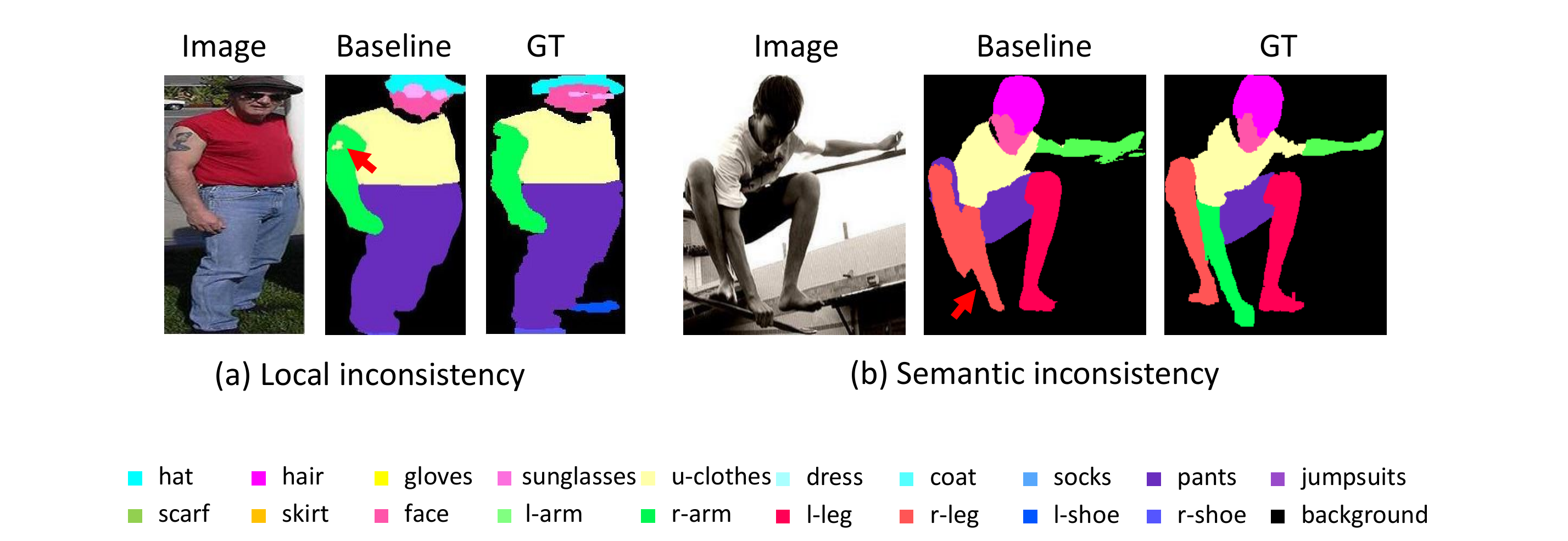}
\caption{Drawbacks of the pixel-wise classification loss. (a) Local inconsistency, which leads to a hole on the arm. (b) Semantic inconsistency, which causes unreasonable human poses. The inconsistencies are indicated by red arrows.}
\label{fig:motivation}
\end{figure}

Based on CNN architecture, the \emph{pixel-wise classification loss} is usually used \cite{liang2015deep,xia2016pose,gong2017look} which punishes the classification error for each pixel. Despite providing an effective baseline, the pixel-wise classification loss which is designed for per-pixel category prediction, has two drawbacks.
First, the pixel-wise classification loss may lead to  \emph{local inconsistency}, such as holes and blur. The reason is that it merely penalizes the false prediction on every pixel without explicitly considering the correlation among the adjacent pixels. For illustration, we train a baseline model (see Section \ref{sec:baseline}) with the pixel-wise classification loss. As shown in Fig. \ref{fig:motivation}(a), some pixels which belongs to ``arm'' are incorrectly predicted as ``upper-clothes'' by the baseline. This is undesirable but is the consequence of local inconsistency of the baseline loss.
Second, pixel-wise classification loss may lead to \emph{semantic inconsistency} in the overall segmentation map, such as unreasonable human poses and incorrect spatial relationship of body parts. Compared to the local inconsistency, the semantic inconsistency is generated from deeper layers. When only looking at a local region, the learned model does not have an overall sense of the topology of body parts. As shown in Fig. \ref{fig:motivation}(b), the ``arm'' is merged with an adjacent ``leg'', indicating incorrect part topology (three legs). Therefore, the pixel-wise classification loss does not explicitly consider the semantic consistency, so that long-range dependency may not be well captured.

In the attempt to address the inconsistency problems, the conditional random fields (CRFs) \cite{krahenbuhl2011efficient} can be employed as a post processing method. However, CRFs usually handle inconsistency in very limited scope (locally) due to the pairwise potentials, and may even generate worse label maps given poor initial segmentation result. As an alternative to CRFs, a recent work proposes the use of adversarial network \cite{luc2016semantic}. Since the adversarial loss assesses whether a label map is real or fake by joint configuration of many label variables, it can enforce higher-level consistency, which cannot be achieved with pairwise terms or the per-pixel classification loss. Now, an increasing number of works adopt the routine of combining the cross entropy loss with an adversarial loss to produce label maps closer to the ground truth \cite{dai2017scan,moeskops2017adversarial,hung2018adversarial}.

\begin{figure}[t]
\centering
\includegraphics[width=1\linewidth]{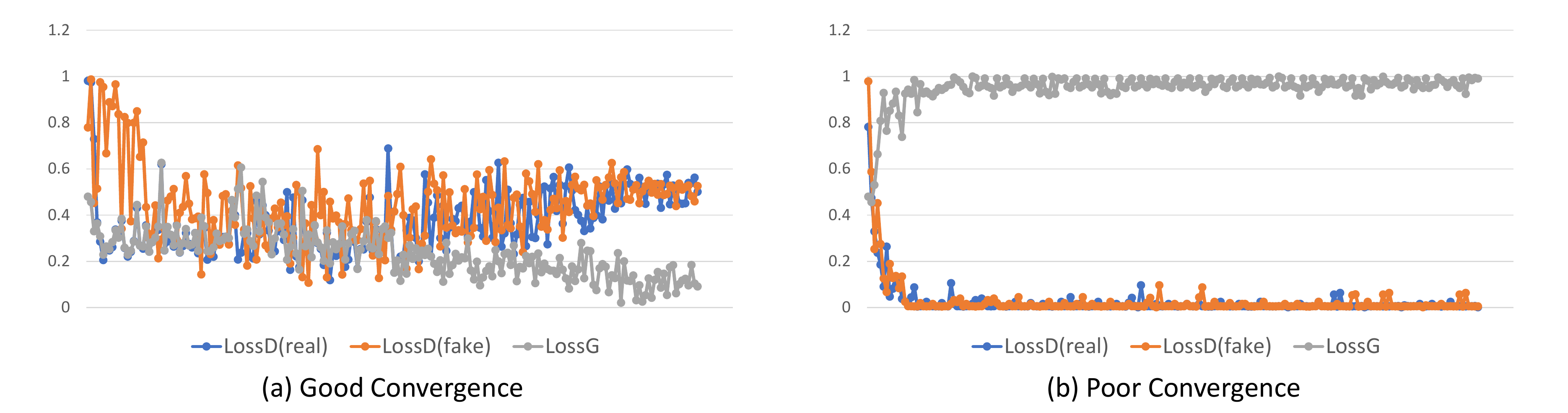}

\caption{Two types of convergence in adversarial network training. $LossD(real)$ and $LossD(fake)$ denote the adversarial losses of discriminator on real and fake image respectively, and $LossG$ denotes the loss of generator. \textbf{(a)} Good convergence, where $LossD(real)$ and $LossD(fake)$ converge to 0.5 and $LossG$ converges to 0. It indicates a successful adversarial network training, where $G$ is able to fool $D$. \textbf{(b)} Poor convergence, where $LossD(real)$ and $LossD(fake)$ converge to 0 and $LossG$ converges to 1. It stands for an unbalanced adversarial network training, where $D$ can easily distinguish generated images from real images.}
\label{fig:poor}
\end{figure}

Nevertheless, the previous adversarial network also has its limitations. First, the single discriminator back propagates only one adversarial loss to the generator. However, the local inconsistency is generated from top layers and the semantic inconsistency is generated from deep layers. The two targeted layers can not be discretely trained with only one adversarial loss. Second, a single discriminator has to look at overall high-resolution image (or a large part of it) in order to supervise the global consistency. As mentioned by numbers of literatures~\cite{denton2015pyramid,karras2017progressive}, it is very difficult for a generator to fool the discriminator on a high-resolution image. As a result, the single discriminator back propagates a maximum adversarial loss invariably, which makes the training unbalanced. We call it \emph{poor convergence problem}, as shown in Fig.~\ref{fig:poor}.

\begin{figure}[t]
\centering
\includegraphics[height=6.5cm]{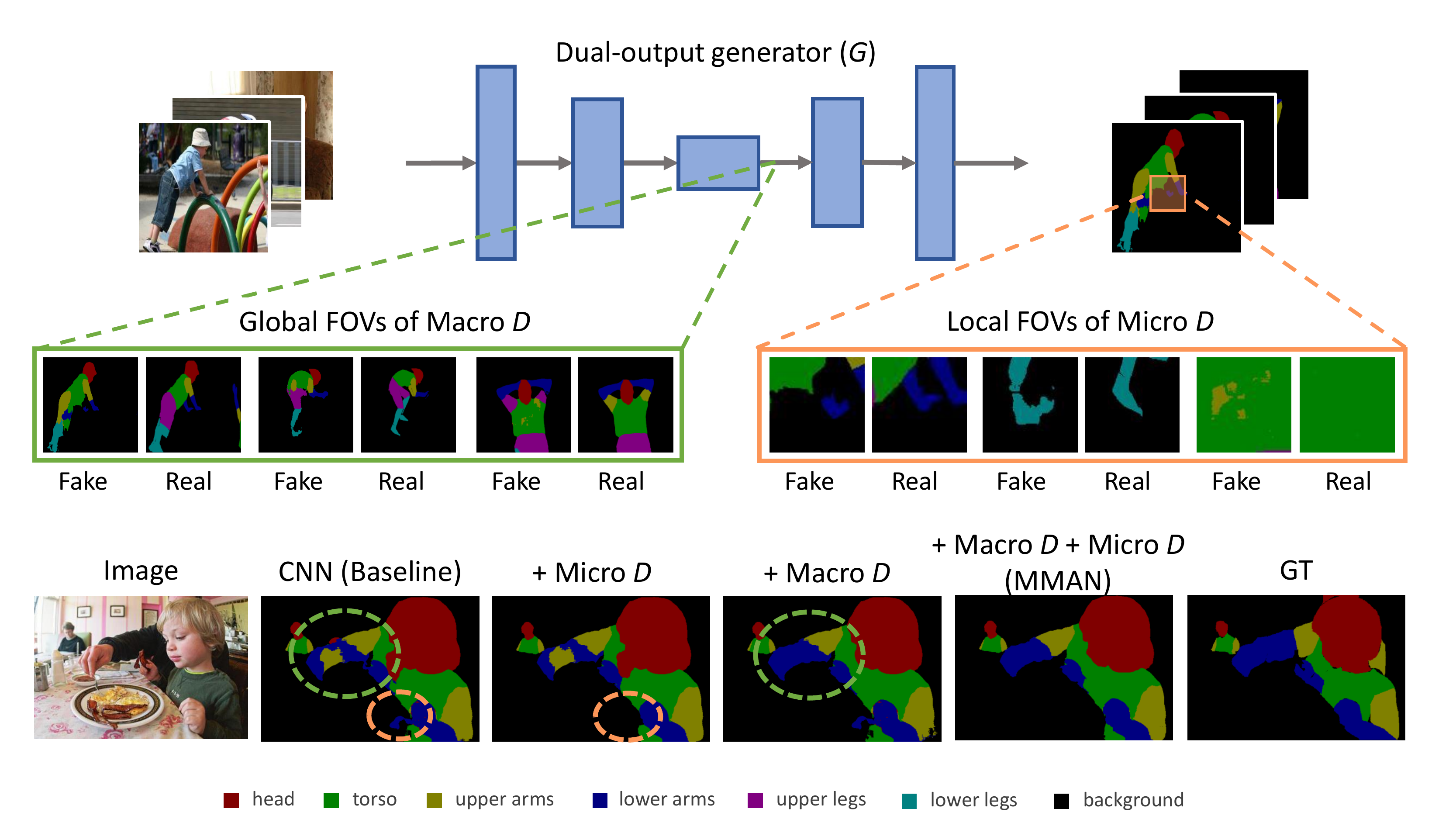}
\caption{\textbf{Top}: A brief pipeline of MMAN. Two discriminators are attached to a CNN-based generator ($G$). The Macro $D$ works on the low-resolution label map and has a global receptive field, focusing on semantic consistency. Micro $D$ focuses on multiple patches and has small receptive fields on high-resolution label map, thus supervising the local consistency. The Macro (Micro) discriminator yields ``fake'' if semantic (local) inconsistency is observed, otherwise it gives ``real''. \textbf{Bottom}:  qualitative results of using Macro $D$, Micro $D$ and MMAN, respectively. We observe that Macro $D$ and Micro $D$ correct semantic inconsistency (\textcolor{green}{green} dashed circle) and local inconsistency (\textcolor{red}{orange} dashed circle), respectively, and that MMAN possesses the merits of both.}
\label{fig:Brief}
\end{figure}

In this paper, the basic objective is to improve the local and semantic consistency of label maps in human parsing. We adopt the idea of adversarial training and at the same time aim to addresses its limitations, \emph{i.e.,} the inferior ability in improving parsing consistency with a single adversarial loss and the poor convergence problem. Specifically, we introduce the Macro-Micro Adversarial Nets (MMAN).
MMAN consists of a dual-output generator ($G$) and two discriminators ($D$), named Macro $D$ and Micro $D$. The three modules constitute two adversarial networks (Macro $AN$, Micro $AN$), addressing the semantic consistency and the local consistency, respectively. Given an input human image, the CNN-based generator outputs two segmentation maps with different resolution levels, \emph{i.e.,} low resolution and high resolution. The input of Macro ${D}$ is a low-resolution segmentation map, and the output is the confidence score of semantic consistency. The input of Micro $D$ is the high-resolution segmentation result, and its outputs is the confidence score of local consistency. A brief pipeline of the proposed framework is shown in Fig.~\ref{fig:Brief}.
It is in two critical aspects that MMAN departs from previous works.
First, our method explicitly copes with the local inconsistency and semantic inconsistency problem using two task-specific adversarial networks individually. Second, our method does not use large-sized FOVs on high-resolution image, so we can avoid the poor convergence problem. More detailed description of the merits of the proposed network is provided in Section \ref{sec:merits}.

Our contributions are summarized as follows:
\begin{itemize}
\item We propose a new framework called Macro-Micro Adversarial Network (MMAN) for human parsing. The Macro $AN$ and Micro $AN$ focus on semantic and local inconsistency respectively, and work in complementary way to improve the parsing quality.

\item The two discriminators in our framework achieve local and global supervision on the label maps with small field of views (FOVs), which avoids the poor convergence problem caused by high-resolution images.

\item The proposed adversarial net achieves very competitive mIoU on the LIP and PASCAL-Person-Part datasets, and can be well generalized on a relatively small dataset PPSS.
\end{itemize}

\section{Related works}
Our review focuses on three lines of literature most relevant to our work, \emph{i.e.,} CNN-based human parsing, the conditional random fields (CRFs) and the adversarial networks.

\textbf{Human parsing.}
Recent progress in human parsing has been due to the two factors: 1) the available of the large-scale datasets \cite{gong2017look,liang2015deep,luo2013pedestrian,chen2014detect}. Comparing to the small datasets, the large-scale datasets contain the common visual variance of people and provide a comprehensive evaluation.
2) the end-to-end learned model. Human parsing demands understanding the person on the pixel level. The recent works apply the convolutional neural network (CNN) to learn the segmentation result in an end-to-end manner. In~\cite{xia2016pose}, human poses are extracted in advance and utilized as strong structural cues to guide the parsing. In~\cite{liang2015human}, four human-related contexts are integrated into a unified network. A novel human-related grammar is presented by \cite{park2017attribute} which infers human body pose and human part segmentation jointly.

\textbf{Conditional random fields}
Using the pixel-wise classification loss, CNN usually ignores the micro context between pixels and the macro context between semantic parts.  Conditional random fields (CRFs)~\cite{krahenbuhl2011efficient,liu2015semantic,li2017holistic} are one of the common methods to enforce spatial contiguity in the output label maps. Served as a post-process procedure for image segmentation, CRFs further fine-tune the output map. However, the most common used CRFs are with pair-wise potentials \cite{chen2016deeplab,luo2016accurate}, which has very limited parameters and handles low-level inconsistencies with a small scope. Higher-order potentials \cite{kohli2009robust,li2017holistic} have also been observed to be effective in enforcing the semantic validity, but the corresponding energy pattern and the clique form are usually difficult to design. In summary, the utilization of context in CNN remains an open problem.

\textbf{Adversarial networks.}
Adversarial networks have demonstrated the effectiveness in image synthesis \cite{isola2017image,odena2016conditional,reed2016learning,zhong2018camera,zhong2018generalizing}. By minimizing the adversarial loss, the discriminator leads the generator to produce high-fidelity images. In \cite{luc2016semantic}, Luc \emph{et al.} add the adversarial loss for training semantic segmentation and yield the competitive results. Similar idea then has been applied in street scene segmentation~\cite{hung2018adversarial} and medical image segmentation~\cite{dai2017scan,moeskops2017adversarial}.
Contemporarily, an increasing body of literature \cite{denton2015pyramid,karras2017progressive} report the difficulty of training the adversarial networks on the high-resolution images. Discriminator can easily recognize the fake high-resolution image, which leads to the training unbalance. The generator and discriminator are prone to stuck in a local minimum.

The main difference between MMAN and the adversarial learning methods above is that the we explicitly endow adversarial training with the macro and micro subtasks. We observe that the two subtasks are complementary to each other to achieve superior parsing accuracy to the baseline with a single adversarial loss and are able to reduce the risk of the training unbalance.

\section{Macro-Micro Adversarial Network}
\begin{figure}[t]
\centering
\includegraphics[height=6.5cm]{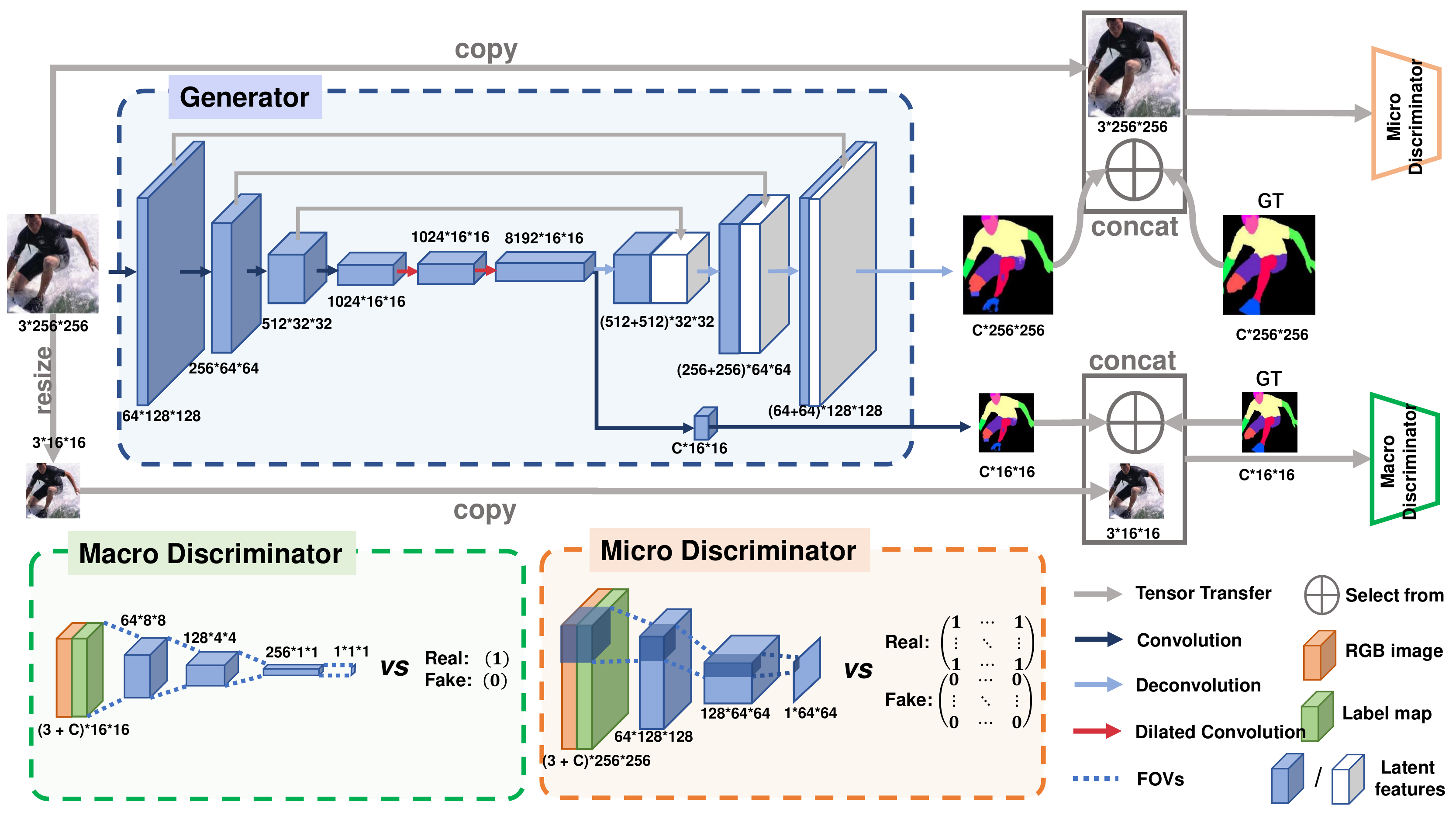}
\caption{MMAN has three components: a dual-output generator (\textcolor{blue}{blue} dashed box), a Macro discriminator (\textcolor{green}{green} dashed box) and a Micro discriminator (\textcolor{red}{orange} dashed box). Given an input image of size $3\times256\times256$, the generator $G$ first produces a low-resolution ($8192\times16\times16$) tensor, from which a low-resolution label map ($C\times16\times16$) and a high-resolution label map ($C\times256\times256$) are generated, where $C$ is the number of classes.
Finally, for the each label map (sized $C\times16\times16$, for example), we concatenate it with an RGB image (sized $3\times16\times16$) along the 1st axis (number of channels), which is fed into the corresponding discriminator.}
\label{fig:Framework}
\end{figure}

Figure \ref{fig:Framework} illustrates the architecture of the proposed Macro-Micro Adversarial Network. The network consists of three components, \emph{i.e.,} a dual-output generator ($G$) and two task-specific discriminators ($D_{Ma}$ and $D_{Mi}$). Given an input image of size $3\times256\times 256$, $G$ outputs two label maps of size $C\times 16\times 16$ and $C\times256\times256$, respectively. $D_{Ma}$ supervises the entire label map of $C\times 16\times 16$ and $D_{Mi}$ focuses on patches of the label map of size $C\times256\times256$, respectively, so that global and local inconsistencies are penalized. In Section \ref{sec:obj}, we illustrate the training objectives, followed by the structure illustration in Section \ref{sec:baseline}, \ref{sec:ma} and \ref{sec:mi}. The merits of the proposed network are discussed in Section \ref{sec:merits}.

\subsection{Training Objectives} \label{sec:obj}
Given a  human image $x$ of shape $3\times H\times W$  and a target label map $y$ of shape $C\times H\times W$ where $C$ is the number of classes including the background, the traditional pixel-wise classification loss (multi-class cross-entropy loss) can be formulated as:
\begin{align}
	\mathcal{L}_{mce}(G) = \sum_{i=1}^{H\times W}\sum_{c=1}^{C} - y_{ic}\log{\hat{y}_{ic}},\label{mce}
\end{align}
where $\hat{y}_{ic}$ denotes the predicted probability of the class $c$ on the $i$-th pixel. The $y_{ic}$ denotes the ground truth probability of the class $c$ on the $i$-th pixel. If the $i$-th pixel belongs to class $c$, $y_{ic}=1,$ else $y_{ic}=0$.

To enforce the spatial consistency, we combine the pixel-wise classification loss with the adversarial loss. It can be formulated as:
\begin{align}
    \mathcal{L}_{mix}(G, D) = \mathcal{L}_{mce}(G) + \lambda \mathcal{L}_{adver}(G, D),\label{mixed}
\end{align}
where $\lambda$ controls the relative importance of the pixel-wise classification loss and the adversarial loss. Specifically, the adversarial loss $\mathcal{L}_{adver}(G, D)$ is:
\begin{align}
    \mathcal{L}_{adver}(G,D) = &\mathbb{E}_{x,y}[\log D(x,y)] + \nonumber \\
                 &\mathbb{E}_{x}[\log (1-D(x,G(x))].\label{cGAN_equation}
\end{align}

As shown in Fig. \ref{fig:Framework}, the proposed MMAN employs the \emph{``cross-entropy loss + adversarial loss''} to supervise both the bottom and top output from the generator $G$:
\begin{align}
    \mathcal{L}_{MMAN}(G, &D_{Ma}, D_{Mi}) = \mathcal{L}_{adver}(G,D_{Ma}) + \lambda_{1} \mathcal{L}_{mce_{l}}(G) \;+ \nonumber \\
     &\lambda_{2} \mathcal{L}_{adver}(G,D_{Mi}) + \lambda_{3} \mathcal{L}_{mce_{h}}(G),\label{eq:total}
\end{align}
where $\mathcal{L}_{mce_l}(G)$ donates the cross-entropy loss between the low-resolution output and the small-sized target label map, while the $\mathcal{L}_{mce_h}(G)$ refers to the cross-entropy loss between the high-resolution output and the original ground-truth label map.
Similarly, $\mathcal{L}_{adver}(G,D_{Ma})$ is the adversarial loss focusing on the low-resolution map, and $\mathcal{L}_{adver}(G,D_{Mi})$ is based on the high-resolution map.
The hyper parameters $\lambda_{1}$, $\lambda_{2}$ and $\lambda_{3}$  control the relative importance of the four losses. The training objective of MMAN is:

\begin{align}
G^*, D_{Ma}^*, D_{Mi}^* = &\arg\min_G\max_{\ssmall{D_{Ma}, D_{Ma}}} \mathcal{L}_{MMAN}(G, D_{Ma}, D_{Mi}).\label{eq:objective}
\end{align}

We solve Eq.~\ref{eq:objective} by alternate between optimizing $G$, $D_{Ma}$ and $D_{Mi}$ until $\mathcal{L}_{MMAN}(G, D_{Ma}, D_{Mi})$ converges.

\subsection{Dual-output Generator}\label{sec:baseline}
For the generator ($G$), we utilize DeepLab-ASPP \cite{chen2016deeplab} framework with ResNet-101 \cite{he2016deep} model pre-trained on the ImageNet dataset \cite{deng2009imagenet} as our starting point due to its simplicity and effectiveness. We augment DeepLab-ASPP architecture with cascaded upsampling layers and skip connect them with early layers, which is similar with U-net~\cite{ronneberger2015u}. Furthermore, we add a bypass to output the deep feature tensor from the bottom layers and transfer it to a label map with a convolution layer. The small-sized label map severs as the second output in parallel with the original sized label map from the top layer. We refer to the augmented dual-output architecture as Do-DeepLab-ASPP and adopt it as our baseline. For the dual output, we supervise the cross-entropy loss from top layers with ground truth label maps of original size, since it can retain visual details. Besides, we supervise the cross-entropy loss of bottom layers with a resized label map, \emph{i.e.,} 1/16 times of the original size. The shrunken label map pays more attentions to the coarse-grained human structure. The same strategy is applied to adversarial loss. We concatenated the respect label map with RGB image of corresponding size along class channel as a strong condition to discriminators.

\subsection{Macro Discriminator} \label{sec:ma}
Macro discriminator ($D_{Ma}$) aims to lead the generator to produce realistic label map that consist with high-level human characteristics, such as reasonable human poses and correct spatial relationship of body parts. $D_{Ma}$ is attached to the bottom layer of $G$ and focuses on an overall low-resolution label map. It consists of 4 convolution layers with kernel size of $4\times4$ and stride of 2. Each convolution layer follows by one instance-norm layer and one LeakyRelu function. Given a output label map from $G$, $D_{Ma}$ downsamples it to $1\times1$ to achieve the global supervision on it. The output of $D_{Ma}$ is the confidence score of semantic consistency.

\subsection{Micro Discriminator} \label{sec:mi}
Micro discriminator ($D_{Mi}$) is designed to enforce the local consistency in label maps. We follow the idea of ``PatchGAN''~\cite{isola2017image} in designing the $D_{Mi}$. Different from $D_{Ma}$ that has a global receptive field on the (shrunken) label map, $D_{Mi}$ only penalizes local error at the scale of image patches. The kernel size of $D_{Mi}$ is $4\times4$ and the stride is 2. Micro $D$ has a shallow structure of 3 convolution layers, each convolution layer follows by one instance-norm layer and one LeakyRelu function. $D_{Mi}$ aims to classify if each $22 \times 22$ patch in an high-resolution image is real or fake, which is suitable for enforcing the local consistency. After running $D_{Mi}$ convolutationally across the label map, we will obtain multiple response from every receptive field. We finally averages all responses to provide the ultimate output of $D_{Mi}$.

\subsection{Discussions}\label{sec:merits}
In CNN-based human parsing, convolution layers go deep to extract part-level features, and deconvolution layers bring the in-depth features back to pixel-level locations. It seems intuitive to arrange the Macro $D$ to deeper layers to supervise high-level semantic features and Micro $D$ to top layers, focusing on low-level visual features. Besides the intuitive motivation, however, we can benefit more from such arrangement. The merits of MMAN are summarized in four aspects.

\textbf{Functional specialization of Macro $D$ and Micro $D$.} Compared with the single discriminator which attempts to solve two levels of inconsistency alone, Macro $D$ and Micro $D$ are specified in addressing one of the two consistency problems. Take Macro $D$ as an example. First, Macro $D$ is attached to the deep layer of $G$. Because the semantic inconsistency is originally generated from the deep layers, a such designed Macro $D$ allows the loss to back propagated to $G$ more directly. Second, Macro $D$ acts on a low-resolution label map that retains the semantic-level human structure while filtering out the pixel-level details. It enforces Macro $D$ to focus on the global inconsistency without disturbing by local errors. The same reasoning applies to Micro $D$. In section~\ref{Variant}, we validate that MMAN consistently outperforms the adversarial networks with a single adversarial loss \cite{luc2016semantic,dai2017scan}.

\textbf{Functional complementarity of Macro $D$ and Micro $D$.} As mentioned in \cite{xue2017segan}, supervising classification loss in early deep layers can offer a good coarse-grained initialization for later top layers. Correspondingly, decreasing the loss in top layers can remedy the coarse semantic feature with fine-grained visual details. We assume that the adversarial loss has the same characteristic to work in complementary pattern. We clarify our hypothesis in Section~\ref{Ablation}.

\textbf{Small FOVs to avoid poor convergence problem.} Reported by increasing literatures \cite{denton2015pyramid,karras2017progressive}, the existing adversarial networks have drawbacks in coping with complex high-resolution images. In our framework, Macro $D$ acts on a low-resolution label map and Micro $D$ has multiple but small FOVs on a high-resolution label map. As a result, both Macro $D$ and Micro $D$ avoid using large FOVs as the actual input, which effectively reduce the convergence risk caused by high resolution. We show this benefit in Section~\ref{Variant}.

\textbf{Efficiency.} Comparing with the single adversarial network~\cite{luc2016semantic,dai2017scan}, MMAN achieves the supervision across the overall images with two shallower discriminators, which have fewer parameters. It also owning to the small FOVs of the discriminators. The efficiency of MMAN is showed in variant study in Section~\ref{Variant}.

\section{Experiment}

\subsection{Dataset}
\textbf{LIP}~\cite{gong2017look} is a recently introduced large-scale dataset, challenging in the severe pose complexity, heavy occlusions and body truncation. It contains 50,462 images in total, including 30,362 for training, 10,000 for testing and 10,000 for validation. LIP defines 19 human part (clothes) labels, including hat, hair, sunglasses, upper-clothes, dress, coat, socks, pants, gloves, scarf, skirt, jumpsuits, face, right arm, left arm, right leg, left leg, right shoe and left shoe, and  a background class.

\textbf{PASCAL-Person-Part}~\cite{chen2014detect} annotates the human part segmentation labels and is a subset of PASCAL-VOC 2010~\cite{pascal-voc-2010}. PASCAL-Person-Part includes 1,716 images for training and 1,817 for testing. In this dataset, an image may contain  multiple persons with unconstrained poses and environment. Six human body part classes and the background class are annotated.

\textbf{PPSS}~\cite{luo2013pedestrian} includes 3,673 annotated samples, which are divided into a training set of 1,781 images and a testing set of 1,892 images. It defines seven human parts and a background class. Collected from 171 surveillance videos, the dataset can reflect the occlusion and illumination variation in real scene.

\textbf{Evaluation metric.}
The human parsing accuracy of each class is measured in terms of pixel intersection-over-union (IoU). The mean intersection-over-union (mIoU) is computed by averaging the IoU across all classes. We use both IoU for each class and mIoU as evaluation metrics for each dataset.

\subsection{Implementation Details}
In our implementation, input images are resized so that its shorter side is fixed to 288. A $256\times256$ crop is randomly sampled from the image or its horizontal flipped version. The per-pixel mean is subtracted from the cropped image. We adopt instance normalization~\cite{instancenorm} after each convolution. For the hyperparameters in Eq.\ref{eq:total}, we set $\lambda_{1} = 25$, $\lambda_{2} = 1$ and $\lambda_{3} = 100$. For the down-sampling network of the generator, we use the ImageNet~\cite{deng2009imagenet} pretrained network as initialization. The weights of the rest of the network are initialized from scratch using Gaussian distribution with standard deviation as 0.001. We use Adam optimizer \cite{kingma2014adam} with a mini-batch size of 1. We set $\beta1 = 0.9$, $\beta2 = 0.999$ and $weight decay = 0.0001$. Learning rate starts from 0.0002. On the LIP dataset, learning rate is divided by 10 after 15 epochs, and the models are trained for 30 epochs. On the Pascal-Person-Part dataset, learning rate is divided by 10 after 25 epochs, and the models are trained for 50 epochs. We use dropout in the deconvolution layers, following the practice in~\cite{isola2017image}. We alternately optimize the $D$ and $G$.
During testing, we average the per-pixel classification scores at multiple scales, \emph{i.e.,} testing images are resized to \{0.8, 1, 1.2\} times of their original size.

\subsection{Comparison with the State-of-the-Art Methods}
In this section, we compare our result with the state-of-the-art methods on the three datasets. First, on the \textbf{LIP dataset}, we compare MMAN with five state-of-the-art methods in Table \ref{tab:LIP}. The proposed MMAN yields an mIoU of 46.65\%, while the mIoU of the five competing methods is 18.17\%~\cite{badrinarayanan2017segnet}, 28.29\%~\cite{long2015fully}, 42.92\%~\cite{chen2016attention}, 44.13\%~\cite{chen2016deeplab} and 44.73\%~\cite{gong2017look}, respectively. For a fair comparison, we further implement ASN~\cite{luc2016semantic} and SSL~\cite{gong2017look} on our baseline, \emph{i.e}, Do-Deeplab-ASPP. On the same baseline, MMAN outperforms ASN~\cite{luc2016semantic} and SSL~\cite{gong2017look} by +1.40\% and +0.62\% in terms of mIoU, respectively. It clearly indicates that our method outperforms  the state of the art. The comparison of per-class IoU indicates that improvement is mainly from classes which are closely related to human pose, such as arms, legs and shoes. In particular, MMAN is capable of distinguishing between ``left'' and ``right'', which gives a huge boost in following human parts: more than +2.5\% improvement in left/right arm, more than +10\% improvement in left/right leg and more than +5\% improvement in left/right shoe. The comparison implies that MMAN is capable of enforcing the consistency of semantic-level features, \emph{i.e.,} human pose.

\begin{table}[t]\ssmall
	\setlength{\tabcolsep}{0.72pt}
    \caption{Method comparison of per-class IoU and mIoU on LIP validation set.}
    \textls{
	\begin{tabular*}{\textwidth}{l|ccccccccccccccccccccc}
    \toprule[0.7pt]
Method & hat & hair & glov & sung & clot & dress & coat & sock & pant & suit & scarf & skirt & face & l-arm & r-arm & l-leg & r-leg & l-sh & r-sh & bkg & avg\\

\hline

\tiny SegNet\cite{badrinarayanan2017segnet} & \tiny 26.60 & \tiny 44.01 & \tiny 0.01  & \tiny 0.00  & \tiny 34.46 & \tiny 0.00  & \tiny 15.97 & \tiny 3.59  & \tiny 33.56 & \tiny 0.01  & \tiny 0.00  & \tiny 0.00  & \tiny 52.38 & \tiny 15.30 & \tiny 24.23 & \tiny 13.82 & \tiny 13.17 & \tiny 9.26  & \tiny 6.47  & \tiny 70.62 & \tiny 18.17  \\

\tiny FCN-8s\cite{long2015fully} & \tiny 39.79 & \tiny 58.96 & \tiny 5.32  & \tiny 3.08  & \tiny 49.08 & \tiny 12.36 & \tiny 26.82 & \tiny 15.66 & \tiny 49.41 & \tiny 6.48  & \tiny 0.00  & \tiny 2.16  & \tiny 62.65 & \tiny 29.78 & \tiny 36.63 & \tiny 28.12 & \tiny 26.05 & \tiny 17.76 & \tiny 17.70 & \tiny 78.02 & \tiny 28.29  \\

\tiny Attention\cite{chen2016attention} & \tiny 58.87 & \tiny 66.78 & \tiny 23.32 & \tiny 19.48 & \tiny 63.20 & \tiny 29.63 & \tiny 49.70 & \tiny 35.23 & \tiny 66.04 & \tiny 24.73 & \tiny 12.84 & \tiny 20.41 & \tiny 70.58 & \tiny 50.17 & \tiny 54.03 & \tiny 38.35 & \tiny 37.70 & \tiny 26.20 & \tiny 27.09 & \tiny 84.00 & \tiny 42.92   \\

\tiny DeepLab-ASPP\cite{chen2016deeplab} & \tiny 56.48 & \tiny 65.33 & \tiny 29.98 & \tiny 19.67 & \tiny 62.44 & \tiny 30.33 & \tiny 51.03 & \tiny 40.51 & \tiny 69.00 & \tiny 22.38 & \tiny 11.29 & \tiny 20.56 & \tiny 70.11 & \tiny 49.25 & \tiny 52.88 & \tiny 42.37 & \tiny 35.78 & \tiny 33.81 & \tiny 32.89 & \tiny 84.53 & \tiny 44.03 \\

\tiny Attention+SSL\cite{gong2017look} & \tiny \textcolor{blue}{59.75} & \tiny \textcolor{blue}{67.25} & \tiny 28.95 & \tiny 21.57 & \tiny \textcolor{blue}{65.30} & \tiny 29.49 & \tiny 51.92 & \tiny 38.52 & \tiny 68.02 & \tiny 24.48 & \tiny \textcolor{blue}{14.92} & \tiny \textcolor{blue}{24.32} & \tiny 71.01 & \tiny 52.64 & \tiny 55.79 & \tiny 40.23 & \tiny 38.80 & \tiny 28.08 & \tiny 29.03 & \tiny 84.56 & \tiny 44.73  \\

\hline
\tiny Do-DeepLab-ASPP & \tiny 56.16 & \tiny 65.28 & \tiny 28.53 & \tiny 20.16 & \tiny 62.54 & \tiny 29.04 & \tiny 51.22 & \tiny 38.00 & \tiny 69.82 & \tiny 22.62 & \tiny 10.63 & \tiny 19.94 & \tiny 69.88 & \tiny 51.83 & \tiny 53.01 & \tiny 45.68 & \tiny 46.08 & \tiny 35.82 & \tiny 34.72 & \tiny 83.47 & \tiny 44.72  \\

\tiny Macro AN & \tiny 57.24 & \tiny 65.28 & \tiny 28.87 & \tiny 19.56 & \tiny 64.02 & \tiny 27.51 & \tiny 51.39 & \tiny 38.13 & \tiny 70.11 & \tiny 22.81 & \tiny 9.05 & \tiny 19.35 & \tiny 68.60 & \tiny 54.19 & \tiny 56.29 & \tiny 50.57 & \tiny 51.22 & \tiny 37.15 & \tiny 37.42 & \tiny 83.25 & \tiny 45.60    \\

\tiny Micro AN & \tiny 57.47 & \tiny 65.05 & \tiny 28.66 & \tiny 16.93 & \tiny 63.95 & \tiny \textcolor{blue}{31.45} & \tiny 51.11 & \tiny 39.64 & \tiny 70.85 & \tiny 25.58 & \tiny 6.87 & \tiny 18.96 & \tiny 68.89 & \tiny 53.62 & \tiny 56.69 & \tiny 49.81 & \tiny 49.42 & \tiny 35.35 & \tiny 35.65 & \tiny 84.46 & \tiny 45.52    \\

\tiny ASN~\cite{luc2016semantic}  & \tiny 56.92 & \tiny 64.34 & \tiny 28.07 & \tiny 17.78 & \tiny 64.90 & \tiny 30.85 & \tiny 51.90 & \tiny 39.75 & \tiny \textcolor{blue}{71.78} & \tiny 25.57 & \tiny 7.97 & \tiny 17.63 & \tiny 70.77 & \tiny 53.53 & \tiny 56.70 & \tiny 49.58 & \tiny 48.21 & \tiny 34.57 & \tiny 33.31 & \tiny 84.01 & \tiny 45.41\\

\tiny SSL~\cite{gong2017look}  & \tiny 58.21 & \tiny 67.17 & \tiny \textcolor{blue}{31.20} & \tiny \textcolor{blue}{23.65} & \tiny 63.66 & \tiny 28.31 & \tiny \textcolor{blue}{52.35} & \tiny 39.58 & \tiny 69.40 & \tiny \textcolor{blue}{28.61} & \tiny 13.70 & \tiny 22.52 & \tiny \textcolor{blue}{74.84} & \tiny 52.83 & \tiny 55.67 & \tiny 48.22 & \tiny 47.49 & \tiny 31.80 & \tiny 29.97 & \tiny 84.64 & \tiny 46.19\\

\hline
\tiny MMAN & \tiny 57.66 & \tiny 65.63 & \tiny 30.07 & \tiny 20.02 & \tiny 64.15 & \tiny 28.39 & \tiny 51.98 & \tiny \textcolor{blue}{41.46} & \tiny 71.03 & \tiny 23.61 & \tiny 9.65 & \tiny 23.20 & \tiny 69.54 & \tiny \textcolor{blue}{55.30} & \tiny \textcolor{blue}{58.13} & \tiny \textcolor{blue}{51.90} & \tiny \textcolor{blue}{52.17} & \tiny \textcolor{blue}{38.58} & \tiny \textcolor{blue}{39.05} & \tiny \textcolor{blue}{84.75} & \tiny \textcolor{blue}{46.81}\\
        \bottomrule[0.7pt]
	\end{tabular*}
    }
    \label{tab:LIP}
\end{table}

Second, on \textbf{PASCAL-Person-Part}, the comparison is shown in Table \ref{tab:Pascal}. We apply the same model structure used on the LIP dataset to train the PASCAL-Person-Part dataset. Our model yields an mIoU of 58.45\% on the test set. It is higher than most of the compared methods and is only slightly inferior to  ``Attention+SSL'' \cite{gong2017look} by 0.91\%. This is probably due to the human scale variance in this dataset, which can be addressed by the attention algorithm proposed in \cite{chen2016attention} and applied in \cite{gong2017look}.

Therefore, we add a plug-and-play module to our model, \emph{i.e.,} attention network \cite{chen2016attention}. In particular, we employ multi-scale input and use the attention network to merge the results. The final model ``Attention+MMAN'' improves mIoU to 59.91\%,  which is higher than the current state-of-the-art method \cite{gong2017look} by +0.55\%. When we look into the per-class IoU scores, we have similar observations to the those on LIP. The largest improvement can be observed in arms and legs. The improvement over the state-of-the-art methods \cite{gong2017look,liang2016lgLSTM,chen2016attention} is over +0.6\% in upper arms,  over +1.8\% in lower arms,  over +0.4\% in upper legs and over +0.9\% in lower legs, respectively. The comparisons indicate that our method is very competitive.

\begin{table}[t]
\centering
\scriptsize
\tabcolsep 0.08in
\caption{Performance comparison in terms of per-class IoU with five state-of-the-art methods on the PASCAL-Person-Part test set.}
\label{tab:Pascal}
\begin{tabular}{l|cccccccc}
\toprule[0.7pt]
Method                                  &   head    &   torso   &   u-arms  &   l-arms  &   u-legs  &   l-legs  &   bkg &   avg \\
\hline
Deeplab-ASPP~\cite{chen2016deeplab}		&   81.33	&   60.06   &   41.16   &   40.95   &   37.49   &   32.56   &   92.81   &   55.19   \\
HAZN~\cite{xia2016zoom}					&   80.79   &   59.11   &   43.05   &   42.76   &   38.99   &   34.46   &   93.59   &   56.11   \\
Attention~\cite{chen2016attention}      &   81.47   &   59.06   &   44.15   &   42.50   &   38.28   &   35.62   &   93.65   &   56.39   \\
LG-LSTM~\cite{liang2016lgLSTM} 			&   82.72   &   60.99   &   45.40   &   \textcolor{blue}{47.76} &   42.33   &   37.96   &   88.63   &   57.97   \\
Attention + SSL~\cite{gong2017look}~~~~~		& \textcolor{blue}{83.26}   &   62.40   &   47.80   &   45.58   &   42.32   &   39.48   &   94.68   &   59.36   \\
\hline
Do-Deeplab-ASPP  						&   81.82   &   59.53   &   44.80   &   42.79   &   38.32   &   36.38   &   93.91   &   56.79   \\
Macro AN                                &   82.01	&   61.19  	&   45.24   &   44.30   &   39.73   &   36.75   &   93.89   &   57.58   \\
Micro AN                                &   82.44	&   61.35  	&   44.79   &   43.68   &   38.41   &   36.05   &   93.93   &   57.23   \\
\hline
MMAN                                    &   82.46	&   61.41  	&   46.05   &   45.17   &   40.93   &   38.83   &   94.30   &   58.45   \\
Attention + MMAN						&   82.58   &   \textcolor{blue}{62.83} &   \textcolor{blue}{48.49} &   47.37   &   \textcolor{blue}{42.80}  &   \textcolor{blue}{40.40} &   \textcolor{blue}{94.92} &   \textcolor{blue}{59.91} \\
\bottomrule[0.7pt]
\end{tabular}
\end{table}

Third, we deploy the model trained on LIP to the testing set of the \textbf{PPSS dataset} without any fine-tuning. We aim to evaluate the generalization ability of the proposed model.

To make the labels in the LIP and PPSS datasets consistent, we merge the fine-grained labels of LIP into coarse-grained human part labels defined in PPSS. The evaluation result is reported in Table~\ref{tab:PPSS}. MMAN yields an mIoU of 52.11\%, which significantly outperforms DL~\cite{luo2013pedestrian} DDN~\cite{luo2013pedestrian} and ASN~\cite{luc2016semantic} by +16.9\% , +4.9\% and +1.4\%, respectively. Therefore, when directly tested on another dataset with different image styles, our model still yields good performance.

\begin{table}[t]
\centering
\scriptsize
\tabcolsep 0.12in
\caption{Comparison of human parsing accuracy on the PPSS dataset~\cite{luo2013pedestrian}. Best performance is highlighted in \textcolor{blue}{blue}.}
\label{tab:PPSS}
\begin{tabular}{l|cccccccc}
\toprule[0.7pt]
Method                          &   head    &   face    &   up-cloth    &   arms    &   lo-cloth    &   legs    &   bkg     &   avg \\
\hline
DL~\cite{luo2013pedestrian}     &   22.0    &   29.1    &   57.3        &   10.6    &   46.1        &   12.9    &   68.6    &   35.2    \\
DDN~\cite{luo2013pedestrian}    &   35.5	&   44.1    &   68.4        &   17.0    &   \textcolor{blue}{61.7}  &   \textcolor{blue}{23.8}  &   80.0    &   47.2    \\
ASN~\cite{luc2016semantic}      &   51.7    &   \textcolor{blue}{51.0}  &   65.9    &   \textcolor{blue}{29.5}  &   52.8    &   20.3    &   83.8    &   50.7    \\
MMAN							&   \textcolor{blue}{53.1}  &   50.2    &   \textcolor{blue}{69.0}  &   29.4    &   55.9    &   21.4    &   \textcolor{blue}{85.7}  &   \textcolor{blue}{52.1}  \\

\bottomrule[0.7pt]
\end{tabular}
\end{table}

\begin{figure}[t]
\centering
\includegraphics[width=0.99\linewidth]{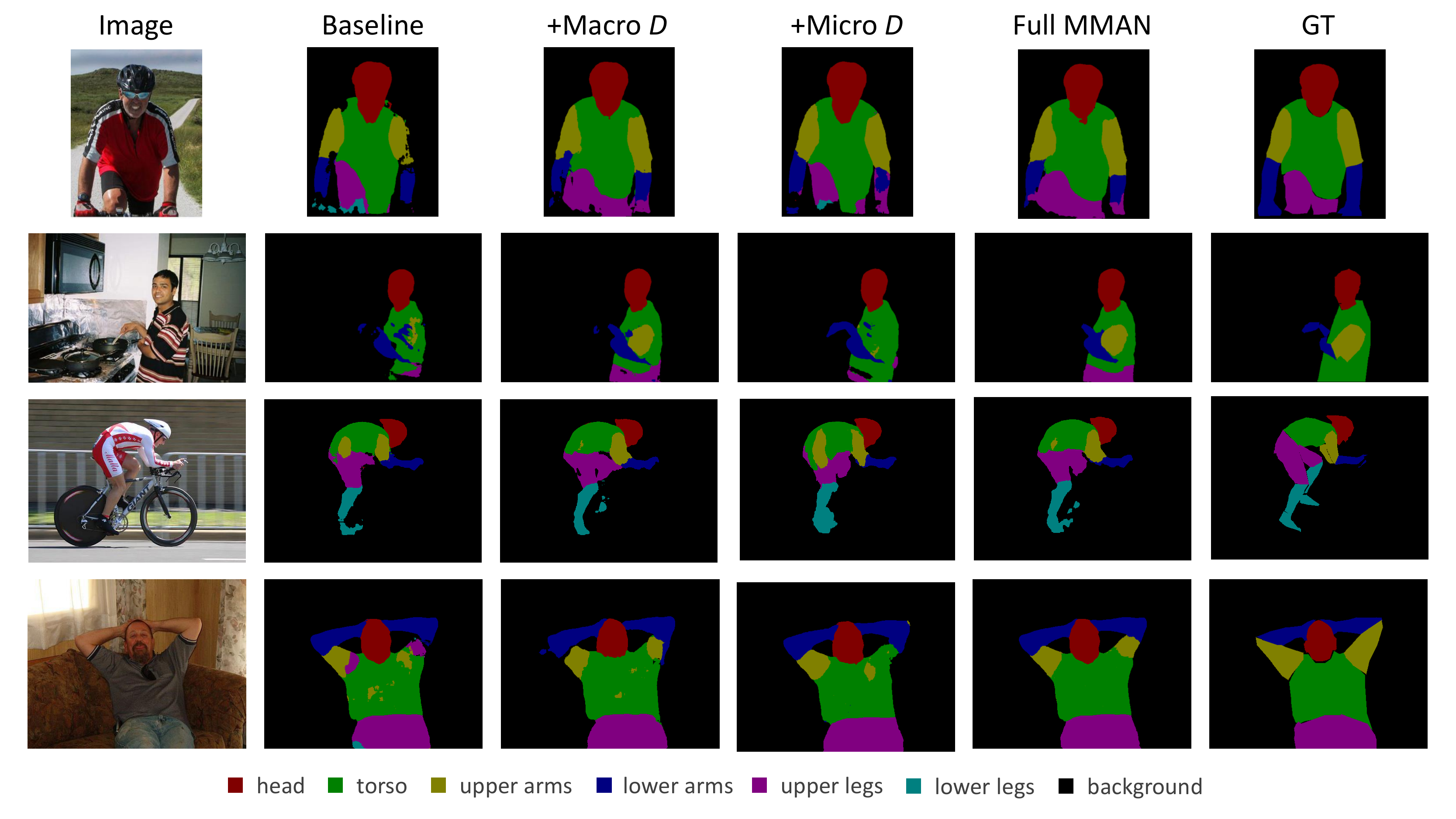}
\caption{Qualitative parsing results on the Pascal-Person-Part dataset.}
\label{fig:result_Pascal}
\vspace{-0.2cm}
\end{figure}

\begin{figure}[!htb]
\centering
\includegraphics[width=0.99\linewidth]{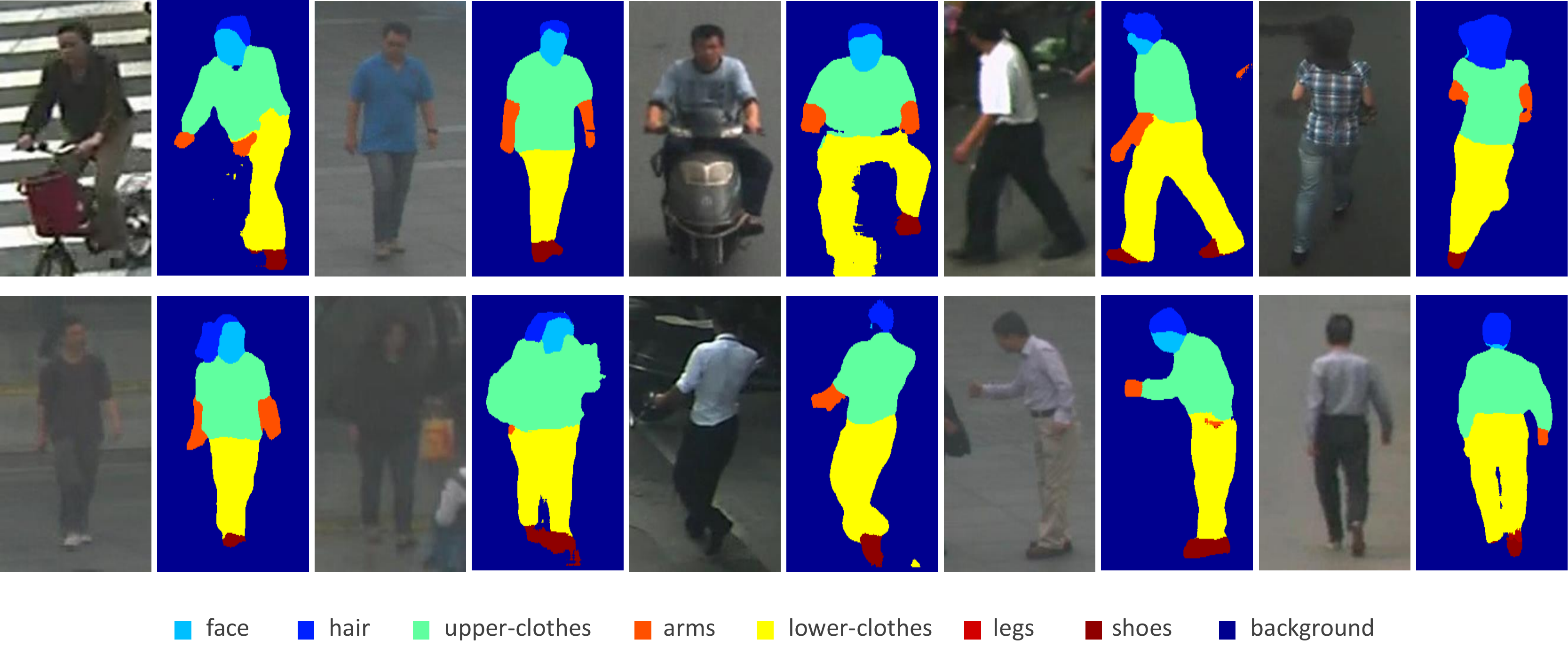}
\caption{Qualitative parsing results on the PPSS dataset. RGB image and the label map are showed in pairs.}
\label{fig:result_PPSS}
\vspace{-0.2cm}
\end{figure}

In Fig.~\ref{fig:result_Pascal}, we provide some segmentation examples obtained by Baseline (Do-Deeplab-ASPP), Baseline+Macro $D$, Baseline+Micro $D$ and full MMAN, respectively. The ground truth label maps are also shown.
We observe that Baseline+Micro $D$ reduces the blur and noise significantly and aids to generate sharp boundaries, and that Baseline+Macro $D$ corrects the unreasonable human poses. The full MMAN method integrates the advantages of both Macro AN and Micro AN and achieves higher parsing accuracy. We also present qualitative results on the PPSS dataset in Fig.~\ref{fig:result_PPSS}.

\subsection{Ablation Study}\label{Ablation}
This section presents ablation studies of our method. Since two components are involved, \emph{i.e.,} Macro $D$ and Micro $D$, we remove them one at a time to evaluate their  contributions respectively. Results on LIP and PASCAL-Person-Part datasets are shown in Table \ref{tab:LIP} and Table \ref{tab:Pascal}, respectively.

On the LIP dataset, when removing Macro $D$ or Micro $D$ from the system, mIoU will drop 1.21\% and 1.29\%, respectively, compared with the full MMAN system. Meanwhile, when compared with the baseline approach, employing Macro $D$ or Micro $D$ alone brings +0.88\% and +0.80\% improvement in mean IoU. Similar observations can be made on the PASCAL-Person-Part dataset as well. 

To further evaluate the respective function of the two different discriminators, we add two external experiments: 1) For Macro $D$, we calculate another mIoU using the low-resolution segmentation maps, which filter out pixel-wise details and retain high-level human structures. So this new mIoU is more suitable for evaluating Macro $D$. 2) For Micro $D$, we count the ``isolated pixels'' in high-resolution segmentation maps, which reflects local inconsistency such as ``holes''. The ``isolated pixel rate'' (IPR) can be viewed as a better indicator for evaluating Micro $D$. We see from Table~\ref{tab:CRF} that Macro $D$ is better than Micro $D$ at improving ``mIoU (low-reso.)'', proving that Macro $D$ \emph{specializes in preserving high-level human structures}. We also see that Micro $D$ is better than Macro $D$ at decreasing IPR, suggesting that Micro $D$ \emph{specializes in improving local consistency} of the result.

\subsection{Variant Study}\label{Variant}
\begin{figure}[t] \label{fig:param_analysis}
\begin{minipage}[c]{.5\linewidth}
\centering
\captionof{table}{Comparison in IPR and mIOUs}
\begin{tabular}{l|c|c|c}
\toprule
\ssmall method & \ssmall IPR & \ssmall mIoU (low-reso.) & \ssmall mIoU (high-reso.) \\
\hline
\ssmall baseline & \ssmall 5.62 & \ssmall 50.66 & \ssmall 44.72\\
\hline
\ssmall +macro $D$ & \ssmall 4.23 & \ssmall 55.79 & \ssmall 45.60\\
\hline
\ssmall +micro $D$ & \ssmall 2.81 & \ssmall53.60 & \ssmall 45.52\\
\hline
\ssmall +CRF & \ssmall \textcolor{blue}{1.53} & \ssmall52.77 & \ssmall 45.45\\
\hline
\ssmall MMAN & \ssmall 2.47 & \ssmall \textcolor{blue}{56.95} & \ssmall \textcolor{blue}{46.81}\\
\bottomrule
\end{tabular}
\label{tab:CRF}
\end{minipage}
\begin{minipage}{.5\linewidth}
\centering
\captionof{table}{Variant study of MMAN.}
\label{table:Variant}
\begin{tabular}{l|c|c|c|c|c|c}
\toprule
\ssmall variant &   \ssmall arch.  &   \ssmall g.FOV     &   \ssmall l.FOV      &   \ssmall \#par    &   \ssmall pc.  &   \ssmall mIoU    \\
\hline
\ssmall sAN   &   \includegraphics[height=0.3cm]{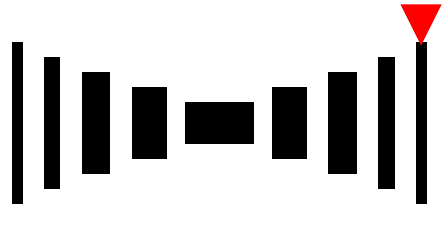}   &   \ssmall $256\times256$  &   \ssmall -                   &   \ssmall 3.2M        &   \ssmall $\surd$     &   \ssmall 45.23       \\
\hline
\ssmall dAN   &   \includegraphics[height=0.3cm]{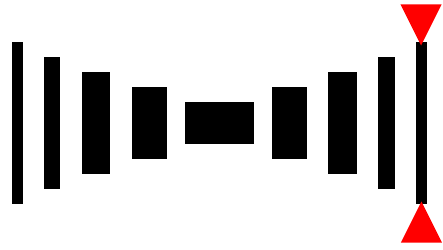}   &   \ssmall $256\times256$  &   \ssmall $22\times22$        &   \ssmall 3.8M        &   \ssmall $\surd$     &   \ssmall 46.15       \\
\hline
\ssmall mAN &   \includegraphics[height=0.3cm]{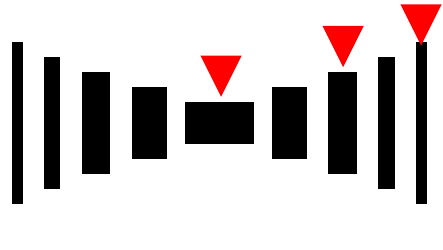}    &   \ssmall $16\times16$    &   \ssmall $22\times22$        &   \ssmall 1.8M        &   \ssmall -           &   \ssmall 46.97       \\
\hline
\ssmall MMAN        &   \includegraphics[height=0.3cm]{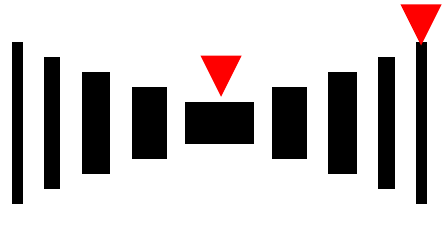}        &   \ssmall $16\times16$    &   \ssmall $22\times22$        &   \ssmall 1.2M        &   \ssmall -           &   \ssmall 46.81       \\
\bottomrule
\end{tabular}
\end{minipage}
\end{figure}

We further evaluate three different variants of MMAN, \emph{i.e.,} Single AN, Double AN, and Multiple AN, on the LIP dataset. Table~\ref{table:Variant} details the numer of parameter, global FOV (g.FOV) and local FOV (l.FOV) sizes, as well as the architecture sketch of each variant. The result of original MMAN is also presented for a clear comparison.

Single AN refers to the traditional adversarial network with only one discriminator. The discriminator is attached to the top layer and has a global receptive field on a $256\times256$ label map. As the result shows, Single AN yields 45.23\% in mean IoU, which is slightly higher than the baseline but lower than MMAN. This result suggests that employing Macro $D$ and Micro $D$ outperforms the single discriminator, which proves the correctness of the analysis in Section \ref{sec:merits}. What is more, we observe the poor convergence (pc) problem when training the Single AN. It is due to the employment of large FOVs on the high-resolution label map.

Double AN has the same number of discriminators with MMAN. The difference lies in that the Double AN attaches the Macro $D$ to the top layer. Compared to Double AN, MMAN significantly improves the result by 0.82\%. The result illustrates the complementary effects of Macro $D$ and Micro $D$: Macro $D$ acts on deep layers and offers a good coarse-grained initialization for later top layers and Micro $D$ helps to remedies the coarse semantic feature with fine-grained visual details. 

Multiple AN is designed to evaluate the parsing accuracy when employing more than two discriminators. To this end, we attach an extra discriminator to the $3rd$ deconvolution layer of $G$. In particular, the discriminator has the same architecture with micro $D$ and focuses on $22\times22$ patches on a $64\times64$ label map. As the result shows in Table~\ref{table:Variant}, employing three discriminators brings very slightly improvement (0.16\%) in mean IoU, but results in more complex architecture and more parameters.

\setlength{\tabcolsep}{1pt}

\section{Conclusions}
In this paper, we introduce a novel Macro-Micro adversarial network (MMAN) for human parsing, which significantly reduces the semantic inconsistency, \emph{e.g.}, misplaced human parts, and the local inconsistency, \emph{e.g.}, blur and holes, in the parsing results. Our model achieves comparative parsing accuracy with the state-of-the-art methods on two challenge human parsing datasets and has a good generalization ability on other datasets. The two adversarial losses are complementary and outperform previous methods that employ a single adversarial loss. Furthermore, MMAN achieves both global and local supervisions with small receptive fields, which effectively avoids the poor convergence problem of adversarial network in handling high-resolution images.
\subsubsection{Acknowledgment.}
This work is partially supported by the National Natural Science Foundation of China (No. 61572211). We acknowledge the Data to Decisions CRC (D2D CRC) and the Cooperative Research Centers Programme for funding this research.

\bibliographystyle{splncs04}
\bibliography{egbib}

\begin{thebibliography}{10}
\providecommand{\url}[1]{\texttt{#1}}
\providecommand{\urlprefix}{URL }
\providecommand{\doi}[1]{https://doi.org/#1}

\bibitem{badrinarayanan2017segnet}
Badrinarayanan, V., Kendall, A., Cipolla, R.: Segnet: A deep convolutional
  encoder-decoder architecture for image segmentation. IEEE transactions on
  pattern analysis and machine intelligence  \textbf{39}(12),  2481--2495
  (2017)

\bibitem{chen2016deeplab}
Chen, L.C., Papandreou, G., Kokkinos, I., Murphy, K., Yuille, A.L.: Deeplab:
  Semantic image segmentation with deep convolutional nets, atrous convolution,
  and fully connected crfs. arXiv preprint arXiv:1606.00915  (2016)

\bibitem{chen2016attention}
Chen, L.C., Yang, Y., Wang, J., Xu, W., Yuille, A.L.: Attention to scale:
  Scale-aware semantic image segmentation. In: Proceedings of the IEEE
  conference on computer vision and pattern recognition. pp. 3640--3649 (2016)

\bibitem{chen2014detect}
Chen, X., Mottaghi, R., Liu, X., Fidler, S., Urtasun, R., Yuille, A.: Detect
  what you can: Detecting and representing objects using holistic models and
  body parts. In: Proceedings of the IEEE Conference on Computer Vision and
  Pattern Recognition. pp. 1971--1978 (2014)

\bibitem{dai2017scan}
Dai, W., Doyle, J., Liang, X., Zhang, H., Dong, N., Li, Y., Xing, E.P.: Scan:
  Structure correcting adversarial network for chest x-rays organ segmentation.
  arXiv preprint arXiv:1703.08770  (2017)

\bibitem{deng2009imagenet}
Deng, J., Dong, W., Socher, R., Li, L.J., Li, K., Fei-Fei, L.: Imagenet: A
  large-scale hierarchical image database. In: Computer Vision and Pattern
  Recognition, 2009. CVPR 2009. IEEE Conference on. pp. 248--255. IEEE (2009)

\bibitem{denton2015pyramid}
Denton, E.L., Chintala, S., Fergus, R., et~al.: Deep generative image models
  using a laplacian pyramid of adversarial networks. In: Advances in neural
  information processing systems. pp. 1486--1494 (2015)

\bibitem{pascal-voc-2010}
Everingham, M., Van~Gool, L., Williams, C.K.I., Winn, J., Zisserman, A.: The
  {PASCAL} {V}isual {O}bject {C}lasses {C}hallenge 2010 {(VOC2010)} {R}esults.
  http://www.pascal-network.org/challenges/VOC/voc2010/workshop/index.html

\bibitem{gan2016concepts}
Gan, C., Lin, M., Yang, Y., de~Melo, G., Hauptmann, A.G.: Concepts not alone:
  Exploring pairwise relationships for zero-shot video activity recognition.
  In: AAAI. p.~3487 (2016)

\bibitem{gong2017look}
Gong, K., Liang, X., Shen, X., Lin, L.: Look into person: Self-supervised
  structure-sensitive learning and a new benchmark for human parsing. arXiv
  preprint arXiv:1703.05446  (2017)

\bibitem{he2016deep}
He, K., Zhang, X., Ren, S., Sun, J.: Deep residual learning for image
  recognition. In: Proceedings of the IEEE conference on computer vision and
  pattern recognition. pp. 770--778 (2016)

\bibitem{hung2018adversarial}
Hung, W.C., Tsai, Y.H., Liou, Y.T., Lin, Y.Y., Yang, M.H.: Adversarial learning
  for semi-supervised semantic segmentation. arXiv preprint arXiv:1802.07934
  (2018)

\bibitem{isola2017image}
Isola, P., Zhu, J.Y., Zhou, T., Efros, A.A.: Image-to-image translation with
  conditional adversarial networks. arXiv preprint  (2017)

\bibitem{karras2017progressive}
Karras, T., Aila, T., Laine, S., Lehtinen, J.: Progressive growing of gans for
  improved quality, stability, and variation. ICLR  (2018)

\bibitem{kingma2014adam}
Kingma, D.P., Ba, J.: Adam: A method for stochastic optimization. arXiv
  preprint arXiv:1412.6980  (2014)

\bibitem{kohli2009robust}
Kohli, P., Torr, P.H., et~al.: Robust higher order potentials for enforcing
  label consistency. International Journal of Computer Vision  \textbf{82}(3),
  302--324 (2009)

\bibitem{krahenbuhl2011efficient}
Kr{\"a}henb{\"u}hl, P., Koltun, V.: Efficient inference in fully connected crfs
  with gaussian edge potentials. In: Advances in neural information processing
  systems. pp. 109--117 (2011)

\bibitem{li2017holistic}
Li, Q., Arnab, A., Torr, P.H.: Holistic, instance-level human parsing. arXiv
  preprint arXiv:1709.03612  (2017)

\bibitem{liang2015deep}
Liang, X., Liu, S., Shen, X., Yang, J., Liu, L., Dong, J., Lin, L., Yan, S.:
  Deep human parsing with active template regression. IEEE transactions on
  pattern analysis and machine intelligence  \textbf{37}(12),  2402--2414
  (2015)

\bibitem{liang2016lgLSTM}
Liang, X., Shen, X., Xiang, D., Feng, J., Lin, L., Yan, S.: Semantic object
  parsing with local-global long short-term memory. In: Proceedings of the IEEE
  Conference on Computer Vision and Pattern Recognition. pp. 3185--3193 (2016)

\bibitem{liang2015human}
Liang, X., Xu, C., Shen, X., Yang, J., Liu, S., Tang, J., Lin, L., Yan, S.:
  Human parsing with contextualized convolutional neural network. In:
  Proceedings of the IEEE International Conference on Computer Vision. pp.
  1386--1394 (2015)

\bibitem{liu2015semantic}
Liu, Z., Li, X., Luo, P., Loy, C.C., Tang, X.: Semantic image segmentation via
  deep parsing network. In: Computer Vision (ICCV), 2015 IEEE International
  Conference on. pp. 1377--1385. IEEE (2015)

\bibitem{long2015fully}
Long, J., Shelhamer, E., Darrell, T.: Fully convolutional networks for semantic
  segmentation. In: Proceedings of the IEEE conference on computer vision and
  pattern recognition. pp. 3431--3440 (2015)

\bibitem{luc2016semantic}
Luc, P., Couprie, C., Chintala, S., Verbeek, J.: Semantic segmentation using
  adversarial networks. arXiv preprint arXiv:1611.08408  (2016)

\bibitem{luo2013pedestrian}
Luo, P., Wang, X., Tang, X.: Pedestrian parsing via deep decompositional
  network. In: Computer Vision (ICCV), 2013 IEEE International Conference on.
  pp. 2648--2655. IEEE (2013)

\bibitem{luo2016accurate}
Luo, Y., Guan, T., Pan, H., Wang, Y., Yu, J.: Accurate localization for mobile
  device using a multi-planar city model. In: Pattern Recognition (ICPR), 2016
  23rd International Conference on. pp. 3733--3738. IEEE (2016)

\bibitem{moeskops2017adversarial}
Moeskops, P., Veta, M., Lafarge, M.W., Eppenhof, K.A., Pluim, J.P.: Adversarial
  training and dilated convolutions for brain mri segmentation. In: Deep
  Learning in Medical Image Analysis and Multimodal Learning for Clinical
  Decision Support, pp. 56--64. Springer (2017)

\bibitem{odena2016conditional}
Odena, A., Olah, C., Shlens, J.: Conditional image synthesis with auxiliary
  classifier gans. arXiv preprint arXiv:1610.09585  (2016)

\bibitem{park2017attribute}
Park, S., Nie, X., Zhu, S.C.: Attribute and-or grammar for joint parsing of
  human pose, parts and attributes. IEEE transactions on pattern analysis and
  machine intelligence  (2017)

\bibitem{reed2016learning}
Reed, S.E., Akata, Z., Mohan, S., Tenka, S., Schiele, B., Lee, H.: Learning
  what and where to draw. In: Advances in Neural Information Processing
  Systems. pp. 217--225 (2016)

\bibitem{ronneberger2015u}
Ronneberger, O., Fischer, P., Brox, T.: U-net: Convolutional networks for
  biomedical image segmentation. In: International Conference on Medical image
  computing and computer-assisted intervention. pp. 234--241. Springer (2015)

\bibitem{instancenorm}
Ulyanov, D., Vedaldi, A., Lempitsky, V.S.: Instance normalization: The missing
  ingredient for fast stylization. CoRR  \textbf{abs/1607.08022} (2016),
  \url{http://arxiv.org/abs/1607.08022}

\bibitem{xia2016zoom}
Xia, F., Wang, P., Chen, L.C., Yuille, A.L.: Zoom better to see clearer: Human
  and object parsing with hierarchical auto-zoom net. In: European Conference
  on Computer Vision. pp. 648--663. Springer (2016)

\bibitem{xia2016pose}
Xia, F., Zhu, J., Wang, P., Yuille, A.L.: Pose-guided human parsing by an
  and/or graph using pose-context features. In: AAAI. pp. 3632--3640 (2016)

\bibitem{xue2017segan}
Xue, Y., Xu, T., Zhang, H., Long, R., Huang, X.: Segan: Adversarial network
  with multi-scale $ l\_1 $ loss for medical image segmentation. arXiv preprint
  arXiv:1706.01805  (2017)

\bibitem{zhang2018self}
Zhang, X., Kang, G., Wei, Y., Yang, Y., Huang, T.: Self-produced guidance for
  weakly-supervised object localization. In: European Conference on Computer
  Vision. Springer (2018)

\bibitem{zhang2018adversarial}
Zhang, X., Wei, Y., Feng, J., Yang, Y., Huang, T.: Adversarial complementary
  learning for weakly supervised object localization. In: IEEE CVPR (2018)

\bibitem{zhong2018generalizing}
Zhong, Z., Zheng, L., Li, S., Yang, Y.: Generalizing a person retrieval model
  hetero- and homogeneously. In: ECCV (2018)

\bibitem{zhong2018camera}
Zhong, Z., Zheng, L., Zheng, Z., Li, S., Yang, Y.: Camera style adaptation for
  person re-identification. In: CVPR (2018)

\bibitem{iccv2017fashiongan}
Zhu, S., Fidler, S., Urtasun, R., Lin, D., Loy, C.C.: Be your own prada:
  Fashion synthesis with structural coherence. In: International Conference on
  Computer Vision (ICCV) (2017)

\end{thebibliography}

\end{document}